%% file: main_iros24.tex
\documentclass[letterpaper, 10 pt, conference]{ieeeconf}  %

\IEEEoverridecommandlockouts                              %

\usepackage{graphicx} %
\usepackage{xcolor}

\makeatletter
\let\NAT@parse\undefined
\makeatother

\usepackage{hyperref}

\usepackage{amsmath}
\usepackage{outlines}
\usepackage{tensor}
\usepackage[hang,flushmargin,symbol]{footmisc}
\usepackage[capitalize]{cleveref}
\usepackage{flushend}
\crefname{section}{Sec.}{Secs.}
\Crefname{section}{Section}{Sections}

\crefname{figure}{Fig.}{Figs.}
\Crefname{figure}{Figure}{Figures}

\crefname{table}{Tab.}{Tabs.}
\Crefname{table}{Table}{Tables}

\crefname{algorithm}{Algo.}{Algos.}
\Crefname{algorithm}{Algorithm}{Algorithms}

\crefname{subappendix}{App.}{Apps.}
\Crefname{subappendix}{Appendix}{Appendices}

\usepackage{xurl}

\usepackage[font=footnotesize]{caption}
\usepackage[font=footnotesize]{subcaption}
\usepackage{booktabs}
\usepackage{multirow}
\usepackage{multicol}
\usepackage{siunitx}
\usepackage{amssymb}

\usepackage{cite}

\usepackage{pifont}

\usepackage{amsmath}
\usepackage{amssymb}
\usepackage{placeins} %
\usepackage{pageslts} %
\input{assets/variable_definitions}

\input{assets/todo_colors}
\hypersetup{
    colorlinks=true,
    linkcolor=blue,
    filecolor=magenta,    
    citecolor=green,
    urlcolor=blue,
    pdftitle={\ourName: \ourNameFull},
}

\title{\LARGE \bf
\ourName: \ourNameFull
}

\author{Nick Heppert, Max Argus, Tim Welschehold, Thomas Brox, Abhinav Valada%
\thanks{Department of Computer Science, University of Freiburg, Germany.} 
\thanks{This work was funded by the Carl Zeiss Foundation with the ReScaLe project and the German Research Foundation (DFG): 417962828, 401269959.}
\thanks{The authors would like to thank Martin Büchner for helping creating the figures and Vilja Lott for data sanity checking.}
}

\begin{document}

\maketitle
\thispagestyle{empty}
\pagestyle{empty}

\begin{abstract}
\input{sections/00_abstract}
\end{abstract}

\input{sections/01_introduction}
\input{sections/02_related_work}
\input{sections/03_method}
\input{sections/04_experimental_setup}

\input{sections/05_conclusion}

\appendix

\input{sections/06_appendix}

{
\footnotesize
\bibliographystyle{ieeetr}
\bibliography{videoimitator}
}

\end{document}

%% file: assets/variable_definitions.tex
\newcommand{\ourName}{\mbox{DITTO}} %
\newcommand{\ourNameFull}{\textbf{D}emonstration \textbf{I}mitation by \textbf{T}rajectory \textbf{T}ransformati\textbf{o}n}
\newcommand{\supersubscript}[3]{\tensor[^{#1}]{#2}{_{#3}}}
\newcommand{\tf}[2]{
    \supersubscript{{#1}}{\mathbf{T}}{{#2}}
}
\newcommand{\point}[2]{
    \supersubscript{{#1}}{\mathbf{p}}{{#2}}
}

\newcommand{\tblPH}{-}

\newcommand{\interCoefficient}{\alpha}
\newcommand{\interCoefficientScaling}{\sigma}

\newcommand{\realSpace}{\mathbb{R}}

\newcommand{\demo}{d}
\newcommand{\live}{l}
\newcommand{\camera}{c}

\newcommand{\hand}{h}

\newcommand{\object}{o}
\newcommand{\goal}{s}

\newcommand{\recording}{R}
\newcommand{\demoRecording}{\recording_\demo}
\newcommand{\liveRecording}{\recording_\live}

\newcommand{\trajectory}{J}
\newcommand{\liveTrajectory}{\trajectory_\live}
\newcommand{\demoTrajectory}{\trajectory_\demo}
\newcommand{\demoTrajectoryWarped}{
    \tensor[^{\live}]{\trajectory}{_{\demo}}
} %
\newcommand{\demoTrajectoryWarpedObject}{
    \tensor[^{\live, \object}]{\trajectory}{_{\demo}}
}
\newcommand{\demoTrajectoryWarpedGoal}{
    \tensor[^{\live, \goal}]{\trajectory}{_{\demo}}
}

\newcommand{\SEThree}{\ensuremath{SE(3)}}

\newcommand{\grasp}{g}
\newcommand{\allGrasps}{G}

\newcommand{\graspTF}{\tf{\camera}{\grasp}}

\newcommand{\tfDemoLiveObject}{\tensor[^{\demo}]{\mathbf{T}}{_{\live}^{\object}}}
\newcommand{\tfDemoLiveGoal}{\tensor[^{\demo}]{\mathbf{T}}{_{\live}^{\goal}}}

\newcommand{\handPositionCamera}{\point{\camera}{\hand}}
\newcommand{\handPositionCameraLive}{\point{\object}{\camera}^{\live}}
\newcommand{\handPositionObject}{\point{\object}{\hand}}

\newcommand{\timestep}{t}
\newcommand{\allTimesteps}{T}

\newcommand{\allDemoObservations}{O}
\newcommand{\observation}{o}
\newcommand{\demoObservation}{\observation_\demo}
\newcommand{\liveObservation}{\observation_\live}

\newcommand{\demoObservationCropped}{\observation_\demo}
\newcommand{\liveObservationCropped}{\observation_\live}

%% file: assets/todo_colors.tex
\definecolor{red}{RGB}{255, 0, 0}   %
\definecolor{orange}{RGB}{255, 77, 0}   %
\definecolor{green}{RGB}{0, 128, 0}   %
\definecolor{purple}{RGB}{160, 32, 240}   %
\definecolor{lightblue}{RGB}{52, 155, 235}   %
\definecolor{darkmagenta}{RGB}{204, 51, 139} %

%% file: sections/00_abstract.tex
Teaching robots new skills quickly and conveniently is crucial for the broader adoption of robotic systems. In this work, we address the problem of one-shot imitation from a single human demonstration, given by an \mbox{RGB-D} video recording. We propose a two-stage process. In the first stage we extract the demonstration trajectory offline. This entails segmenting manipulated objects and determining their relative motion in relation to secondary objects such as containers. In the online trajectory generation stage, we first \mbox{re-detect} all objects, then warp the demonstration trajectory to the current scene and execute it on the robot.
To complete these steps, our method leverages several ancillary models, including those for segmentation, relative object pose estimation, and grasp prediction.
We systematically evaluate different combinations of correspondence and re-detection methods to validate our design decision across a diverse range of tasks. 
Specifically, we collect and quantitatively test on demonstrations of ten different tasks including pick-and-place tasks as well as articulated object manipulation.
Finally, we perform extensive evaluations on a real robot system to demonstrate the effectiveness and utility of our approach in real-world scenarios. We make the code publicly available at \url{http://ditto.cs.uni-freiburg.de}.

%% file: sections/01_introduction.tex
\section{Introduction}

Humans are remarkably good at learning new motion skills from just a few, or even a single demonstration, given by other humans.
Similarly, a common paradigm to teach a robot is imitation learning~\cite{celemin2022interactive}. Here, a human actively demonstrates a skill to a robot either directly on the robot by teleoperating it \cite{Chisari22}, by kinesthetic teaching~\cite{zhao2022hybrid} or alternatively, by performing the task themselves~\cite{Zimmermann20183}. 
Teleoperating a robot to collect demonstrations is possible with various different input devices \cite{zhu2022viola,Chisari22}. Despite this variety, collecting robot demonstrations remains difficult as these devices typically do not share the morphology of the robot and thus, leads to a significant training phase for the human operator or requires an expert teacher.
To circumvent these problems, kinesthetic teaching is an appealing alternative. 
While this reduces the training time for the operator, it is not always beneficial as the operator has to be present in the scene, which can introduce various challenges such as occlusion or restricting the robot's workspace due to safety constraints.

\begin{figure}
    \centering
    \includegraphics[width=\linewidth]{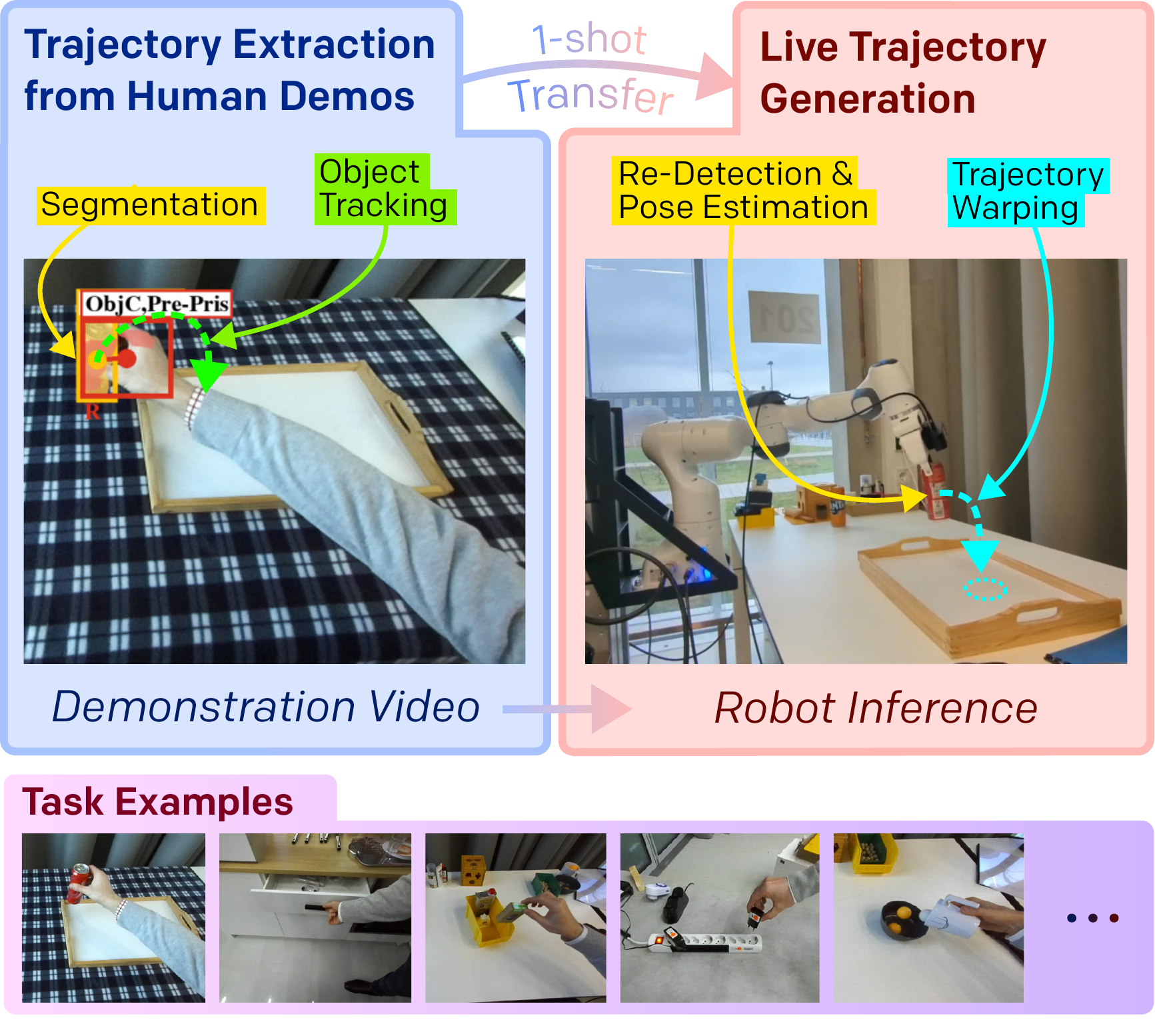}
    \caption{Human demonstration of manipulation actions are transferred to new scenes so that a robot can replicate the manipulation action. For this, we use a two-stage process, first extracting object trajectories by segmenting and tracking objects. Then, we transfer the trajectory to a new scene by re-detection and trajectory transformation according to the re-detected positions. The proposed method is then evaluated on several different tasks.}
    \label{fig:teaser}
\end{figure}

In contrast to these on-robot approaches, in this paper, we propose a novel way to teach robots new tasks by letting them passively observe a human performing a task only once. We move away from end-effector action representations and towards an object-pose-centric perspective~\cite{Vitiello2023} in which we represent the trajectory as a sequence of poses of the object. This allows us to collect demonstration data independent of the embodiment, and we only later perform the transfer from human to robot.

Learning from human demonstrations is preferable over learning from robot demonstrations in many settings as human demonstrations can be collected most naturally and in principle be performed by non-expert users. 
Thus, datasets of humans performing tasks are much easier to collect and more diverse than datasets of robot demonstrations, making them appealing for general learning-based approaches. %
However, learning from human demonstrations introduces additional challenges. The embodiment gap between humans and robots have to be handled with respect to e.g. grasping or kinematic constraints. Furthermore, often there are differences in the visual observation space. Human demonstrations are typically provided from the 3D-person perspective whereas robot demonstrations and skill execution are usually performed from the same perspective. %

We present \ourNameFull~(\ourName{}), illustrated in \cref{fig:teaser}. \ourName{} consists of two stages; first, a trajectory extraction stage in which we leverage recent object-hand-segmenters~\cite{cheng2023towards} and correspondence detectors~\cite{Sun2021} to extract relevant objects and calculate their movement in 3D throughout the demonstration. In the second stage, trajectory generation, we present the robot with a novel scene with the same objects, re-detect the objects, and estimate their relative poses. Based on these poses, we warp and interpolate the trajectory. To eventually execute the task, we use an off-the-shelf grasping method~\cite{Sundermeyer2021} and motion planning algorithms to execute the final trajectory.
We extensively evaluate each phase of our pipeline in an offline procedure and test it on a real robot to examine failure cases. 
We show exemplary produced trajectories in \cref{fig:example_output}.

In summary, we make the following main contributions:
\begin{outline}
    \1 A novel, modular method for 1-shot transfer from \mbox{RGB-D} human manipulation demonstration videos to robots.
    \1 Experiments validating the method and its ablations, conducted both in an offline manner and on a real robot.
    \1 Open source data and code is publicly available at \mbox{\url{http://ditto.cs.uni-freiburg.de}}.
\end{outline}

%% file: sections/02_related_work.tex
\section{Related Work}

Imitation learning is a common paradigm to teach a robot a new task \cite{osa2018algorithmic,celemin2022interactive}. The numerous approaches to this problem can be characterized based on different factors, e.g. by the number of samples used for imitation learning or whether robot or human demonstrations are collected. The following section highlights recent advancements in learning from a few robot demonstrations as well as learning from human demonstrations. 

\subsection{Imitation Learning from Robot Demonstrations}
A typical approach to collect demonstrations is through teleoperating a robot~\cite{Si2021} for example directly through an external controller \cite{zhu2022viola,Chisari22} or a tracked human hand~\cite{qin2022one}. 
Nonetheless, as the human and robot morphology is vastly different, teleoperating a robot can be tedious. Thus, recently, researchers started to investigate how to reduce the amount of needed demonstrations~\cite{Chisari22}.

{\noindent \textit{One- and Few-Shot Robot Imitation:}}
Imitation learning from a single or few robot demonstration is a challenging endeavour as it is difficult to separate the intrinsic geometric invariances that define a task from coincidental ones.
Few-shot methods often make use of sparse representations to learn invariances more efficiently, examples of this include explicit object proposals \cite{zhu2022viola} or keypoint trajectories~\cite{Vecerk2023}. Another strategy is to make use of heterogeneous demonstrations from different tasks~\cite{Yeh2021, Wang2024}. 
One-shot methods often compensate for the availability of other demonstrations by requiring additional inputs such as foreground object segmentation mask~\cite{Argus2020, Vitiello2023, Zhu2023} or demonstrations with singluated objects~\cite{Palo2024}. 

\subsection{Imitation Learning from Human Demonstrations}

While imitation learning from human demonstration videos is compelling%
, human demonstrations suffer from a substantial embodiment shift between humans and robots, even in the case of humanoid robots.
Representation learning from human demonstrations can occur on different levels, starting with visual feature learning, as is done by R3M~\cite{Nair2022}.
In more explicitly structured systems, there are the options of learning visual affordances~\cite{Bahl2023}, value functions~\cite{Bhateja2023} or category-level representations~\cite{Gao2022, wen2022you}. One work made use of eye-in-hand type human demonstration data \cite{Kim2023}.
Similar to \ourName{}, WHIRL~\cite{bahl2022human} extracts a human prior from the given demonstration video by using an off-the-shelf hand pose detector. 
A number of other works also use large numbers of human demonstrations to learn generative models, which generate intermediate representations such as segmentation maps~\cite{Bharadhwaj2023} of hands, flow~\cite{Ko2023}, or trajectories~\cite{Bharadhwaj2023b} on which a policy is based.

{\noindent\textit{One- and few-shot Human Imitation:}}
One and few-shot imitation from human demonstrations is particularly challenging as it compounds the problem of identifying invariances with an embodiment gap. 
Examples of few-shot imitation from human demonstrations include the work by \mbox{Kyriazi~\textit{et al.}~\cite{Kyriazis2015}} and EquivAct~\cite{Yang2023}. Similar to these and our approach, \mbox{Zimmermann~\textit{et al.}~\cite{Zimmermann20183}} make use of human and fiducial marker-based pose estimation to learn manipulation trajectories from few demonstrations.
One-shot imitation from human demonstrations was done by following a meta-learning based approach~\cite{Yu2018}, as well as translating tasks to a shared latent space and generating actions by either using the latent representations as inputs for a reinforcement learning policy~\cite{Pauly2018} or behavioral cloning~\cite{Dasari2020}. Finally, an applied system that explicitly models contacts and computes relative poses was presented by Guo~\textit{et al.}~\cite{Guo2023}.
In contrast, \ourName{} does not require an explicit optimization or learning step and directly transfers to the robot.

\subsection{Segmentation \& Human Action Understanding}
Semantic image segmentation is a well established task in computer vision~\cite{hurtado2022semantic}. Recently, generic segmentation models that segment all objects such as SAM~\cite{Kirillov2023} or segment with few annotated labels such as SPINO~\cite{kappeler2023few} have become widely used. 
Other works, such as Hands23~\cite{Shan2020} focus on human interactions with objects by estimating bounding boxes and segmentation masks for hands, manipulated objects, as well as secondary objects (containers). Other manipulation datasets such as Ego4D~\cite{Grauman2021} have led to the learning of object-centric video representations~\cite{Zhang2023}.

\subsection{Correspondences and Detection}
\label{subsec:related_work:pose_estimation}

The core of \ourName{} strongly relies on a robust relative pose estimation. 
Classicaily, given two unordered point clouds, if correspondences between points are unknown, the iterative closest point (ICP) method can be used to estimate the relative pose between the point clouds. Alternatively, given point correspondences it is possible to directly estimate the relative pose using singular value decomposition \cite{Arun1987}.
Leveraging 
available \mbox{RGB-D} images, these point correspondences can first be obtained via pixel correspondences, for which we highlight a multitude of methods in the following.  

\begin{figure*}[ht!]
    \centering
    \includegraphics[width=\linewidth]{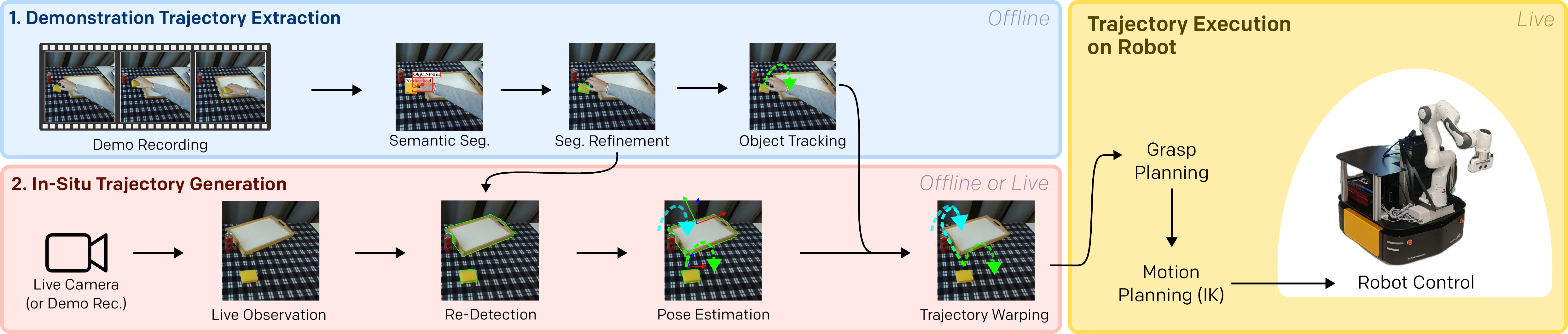}
    \caption{Our method first computes masks and trajectories from demonstration videos, then maps these onto new live observations by accounting for the change in object poses. These warped trajectories can either be evaluated separately or executed on a robot by using grasp planning and IK trajectory solvers. 
    }
    \label{fig:method}
\end{figure*}

There is a wide range of semi-dense correspondence methods available that are typically used for ego-pose estimation problems, e.g. in SLAM. These methods can either be based on analytical features such as SIFT~\cite{Lowe2004} and ORB~\cite{Rublee2011} or learned features such as SuperGlue~\cite{Sarlin2019} and LoFTR~\cite{Sun2021}. 

Dense methods that are used for estimating optical flow, such as RAFT~\cite{Teed2020} or UniMatch~\cite{Xu2023}, can also be used to establish correspondences. Unlike semi-dense methods, flow-prediction methods have the advantage of guaranteeing the existence of correspondences. However, as flow is trained on a data distribution characterized by small rotations and translations, it lacks the capability to establish global correspondences or correspondences between strongly rotated objects.
The classical pose estimation problem is not further discussed here, as it requires known object CAD models. 

Most pose estimation systems rely on upstream detection methods, which aim to identify the rough location of a relevant object.
A simple but yet competitive detection method is CNOS~\cite{Nguyen2023}, which re-detects template images by comparing DINOv2~\cite{Oquab2023} descriptor features that have been pooled according to SAM masks, yielding detections in the form of segmentation masks. This problem is also addressed in other works~\cite{Valassakis2022}.

\subsection{Grasp Generation}

Generating stable grasps based on image or point cloud observations is a well studied problem \cite{newbury2023deep}. Grasp generation methods depend on the gripper geometry. While some methods generate grasps for humanoid hands~\cite{Baek2022, Roa2012}, most relevant to our work, are Contact-GraspNet~\cite{Sundermeyer2021} and Anygrasp~\cite{Fang2022}. Both methods generate grasps for two-finger grippers over whole scenes. Other methods extend these generic setups by making the grasping closed-loop~\cite{Piacenza2023} or by combining it with shape reconstruction~\cite{Chisari2023}. 

%% file: sections/03_method.tex
\section{\ourName{} Method}
\label{sec:method}

Given a demonstration sequence of RGB-D observations $\allDemoObservations = \{ \demoObservation^1, \ldots, \demoObservation^{\allTimesteps} \}$, with length  $\allTimesteps$  and a live observation, $\liveObservation$,  we aim to infer a robot trajectory $\liveTrajectory$ that upon execution will complete the task shown in the demonstration sequence. For this, we take an object-centric approach in which manipulation actions consist of robot end-effector poses that are represented in the manipulated objects frame or a secondary objects frame (e.g. containers).
As outlined in \cref{subsec:method:online}, our method consists of three stages: a demonstration trajectory extraction stage that can be run as offline pre-processing, an in-situ trajectory generation, and trajectory execution, when running online on a robot. We describe these in detail in this section.\looseness=-1

\subsection{Demonstration Trajectory Extraction}
\label{subsec:method:offline}
In the first stage, given a sequence of RGB-D observations, $\allDemoObservations$, we extract the demonstration trajectory $\demoTrajectory$ based on the relative object transformations.

{\noindent \textbf{Object--Hand Segmentation:}} 
\label{subsec:method:object_segmentation}
For all demonstration observations, $\demoObservation^{\timestep} ~\forall~ \timestep$, we use Hands23~\cite{Shan2020} to compute the segmentation masks for the manipulated object and if present, of a secondary object (e.g. a container). While Hands23 provides state-of-the-art results on their benchmark, 
we still occasionally observe suboptimal segmentation masks. This may be due to Hands23 design, which processes a single image rather than a full sequence. However, improved versions of this method are very likely to emerge in the near future and we thus circumvent this problem 
by manually discarding frames for which the segmentation masks are poor, 
leaving more robust object-hand detection to future research. 

{\noindent \textbf{Secondary Object Segmentation}:}
To obtain segmentation masks for the secondary objects, we follow a similar procedure.
We first make use of the secondary object segmentations provided by Hands23.
If these are not present, we fall back onto a heuristic (specified in \cref{sec:appendix:goal_heuristic}) to identify the secondary object mask. The results are again manually verified and bad masks are discarded.

{\noindent \textbf{Object Pose Extraction:}}
Subsequently, given object segmentation masks throughout the sequence, we compute the trajectory of the manipulated object. This is done by performing relative pose estimation between pairs of subsequent time steps $(t, t+1)$. As the translation and rotation between subsequent observations is small and due to hand occlusions the amount of visible object points is low, we prefer methods with a high recall over precision. Thus, from the relative pose estimation methods outlined in \cref{subsec:related_work:pose_estimation}, we chose to estimate correspondences with the flow estimation method RAFT~\cite{Teed2020}.

After establishing correspondences, we filter them based on the object mask. Given the depth images, we then lift the correspondences to 3D and compute the least-squares rigid transformation $\tf{t}{t+1} \in \SEThree$ using singular value decomposition~\cite{Arun1987}. To make this process robust to outliers, we estimate inliers through a RANSAC~\cite{Fischler1981} procedure.
We perform relative pose estimation for all pairs of subsequent images and aggregate them in our demonstration trajectory $\demoTrajectory = \left\{ \tf{1}{2}, \ldots, \tf{\allTimesteps-1}{\allTimesteps} \right\}$.

{\noindent \textbf{Hand Position Extraction:}} 
Lastly, we also estimate the hand position in relation to the manipulated object. 
At the time step where the hand is just about to grasp the object to be manipulated, we lift the center of the 2D hand mask to 3D, resulting in $\handPositionCamera$.  For lifting, we use the given depth image. The hand position relative to the manipulated object $\tf{\camera}{\object}$ is then given by
\begin{equation}
    \handPositionObject = \left( \tf{\camera}{\object} \right)^{-1} \handPositionCamera
\end{equation}
where $\tf{\camera}{\object}$ is a canonical frame of the manipulated object.

\subsection{In-Situ Trajectory Generation}
\label{subsec:method:online}

In the second stage, given a live RGB-D observation $\liveObservation$ with the same objects visible, we will first warp the previously extracted trajectory $\demoTrajectory$ and then execute it. 
Similar to before, we first estimate the relative pose of the manipulated object between the first demonstration observation $\demoObservation^1$ and the live observation $\liveObservation$. If applicable, we do the same for the secondary object. Given the results, we warp the demonstration trajectory $\demoTrajectory$ and retrieve a resulting live trajectory $\liveTrajectory$ which is passed to the robot for execution.

{\noindent\textbf{Re-Detection and Relative-Pose Estimation:}}
While pose estimation systems work on full images, it is common practice to first run detection systems to extract a region containing the relevant objects \cite{Nguyen2023}. This is particularly useful in our case as we use local flow-based correspondence methods. Thus, to improve robustness, we propose to use a modified version of 
CNOS~\cite{Nguyen2023}.\footnote{For simplicity, this modified version is also referred to as CNOS in the paper as the changes are minor.} 

The method originally assumed known CAD models, which are rendered to provide template views for re-detection. We replace the template views with actual views, cropped from demonstration images. Re-detection allows us to create tight crops around the object of interest for both the demonstration observation $\demoObservationCropped$ and the live observation $\liveObservationCropped$.
Similar to \cref{subsec:method:offline}, we perform relative pose estimation on the cropped observations. One drawback of using such a two-step approach is the fact that if the mask re-detection fails the correspondence estimation will also fail as there is no way to retrieve information from the cropped image parts. 

Alternatively, referring to \cref{subsec:related_work:pose_estimation}, we also propose to replace the inherently local flow estimation with a semi-dense, global method, LoFTR~\cite{Sun2021} which does not require an additional detection step. This decision is motivated by the fact that we are faced with the vice-versa case of the previously discussed trajectory extraction. During trajectory generation, we are faced with potentially strong rotations but very little occlusions, thus we can sacrifice a lower recall for higher precision of LoFTR compared to flow-based methods. When using LoFTR we calculate a re-detection mask by fitting a bounding box around all detected correspondences.

Nonetheless, we perform the relative pose estimation step for the manipulated object $\tfDemoLiveObject \in \SEThree$ and if applicable for the secondary object $\tfDemoLiveGoal \in \SEThree$. 

{\noindent \textbf{Trajectory Warping:}}
In the next step, the demonstration trajectory $\demoTrajectory$ is warped to the live scene, yielding the object trajectory in the live scene. 
In the simpler case, with no secondary object present, we use the relative pose change of the object $\tfDemoLiveObject$ and apply it to the demonstration trajectory $\demoTrajectory$ as
\begin{equation}
    \label{eqn:objectframe_transform}
    \demoTrajectoryWarpedObject = \{ \tf{t}{t+1} \tfDemoLiveObject ~~\forall~ \tf{t}{t+1} \in \demoTrajectory\}.
\end{equation}

In the extended case, if a secondary object is present, we perform the same application as in \cref{eqn:objectframe_transform} but with $\tfDemoLiveGoal$
\begin{equation}
        \demoTrajectoryWarpedGoal = \{ \tf{t}{t+1} \tfDemoLiveGoal ~~\forall~ \tf{t}{t+1} \in \demoTrajectory\}.
\end{equation}
We have two potential live trajectories, one based on the object's location $\demoTrajectoryWarpedObject$ and one based on the secondary's object location $\demoTrajectoryWarpedGoal$. To obtain a single final trajectory we smoothly interpolate them\footnote{$\alpha \mathbf{A} \oplus (1-\alpha) \mathbf{A}$ is a generalized addition in the $\SEThree$-space.} using slerp~\cite{Shoemake1985}
\begin{equation}
    \demoTrajectoryWarped = 
    \left\{ 
        \interCoefficient(\timestep) ~ \demoTrajectoryWarpedObject 
        \oplus 
        (1 - \interCoefficient(\timestep)) ~ \demoTrajectoryWarpedGoal
    \right\}
\end{equation}
with Gaussian weights
\begin{equation}
    \interCoefficient(\timestep) = G ( t~|~0,~\interCoefficientScaling (\allTimesteps-1) ) \in \realSpace
\end{equation}
as detailed in \cref{sec:appendix:mixing}.

\subsection{Trajectory Execution on Robot}
Until now we focused on object motion. The next sections describe how to generate robot motion for a specific robot morphology under the assumption of a stable grasp. This addresses the problem of differing embodiments between humans and robots.

{\noindent \textbf{Grasp Generation and Selection}:} 
We use Contact-GraspNet~\cite{Sundermeyer2021} as an off-the-shelf grasping method to detect possible grasps $\allGrasps$ in the live scene. The grasps are computed using the initial live observation and then filtered using the object mask from re-detection to give the subset of grasps only on the object to be manipulated. We further filter grasps via inverse kinematic computation (see below) based on reachability and the potentially resulting full robot trajectory.

Additionally, we use the estimated object pose and the relative hand position to conduct a type of affordance transformation by choosing the grasp with the smallest distance to the hand detection $\graspTF \in \allGrasps$. For this, we leverage the previously estimated relative transformation $\tfDemoLiveObject$ to transform the hand position $\handPositionObject$ back to the live camera frame $\handPositionCameraLive$. 

{\noindent \textbf{Motion Planning and Robot Control:}} 
Based on the grasp, we then compute the end-effector joint trajectory that yields our desired warped object trajectory. We then calculate a full robot joint trajectory for our end-effector pose sequence which includes pre-grasp pose, grasp pose, and all poses of the generated trajectory using KDL kinematics\footnote{Default ROS MoveIt Solver
}. For the execution, we plan and execute the grasp and the generated trajectory separately as we stop in between to close the gripper and confirm the grasp. Note that since the relative pose changes in the generated trajectory are quite small ($\sim0.05~\textrm{[m]}/\sim0.15~\textrm{[rad]}$), the motion planning algorithm is heavily restricted in its search space, potentially inducing failures.\looseness=-1

%% file: sections/04_experimental_setup.tex
\section{Experiments}
We evaluate our approach in three different configurations: through live real robot executions (see \cref{subsec:real_robot_evaluation}) using the robot shown in \cref{fig:robot_setup}, and on offline data
by predicting correspondences (see \cref{subsec:exp:correspondence_tracking}) and the object trajectories (see~\cref{subsec:exp:relative_pose_warping}) of demonstration episodes.
While the offline evaluations have several advantages, the most important are speed of evaluation and the comparability of results, it also remains an inherently approximate evaluation, see \cref{subsec:exp:correspondence_tracking} and \cref{subsec:exp:relative_pose_warping} respectively.

\subsection{Experimental Setup}
\begin{figure}[t]
    \centering
    \includegraphics[width=.8\linewidth]{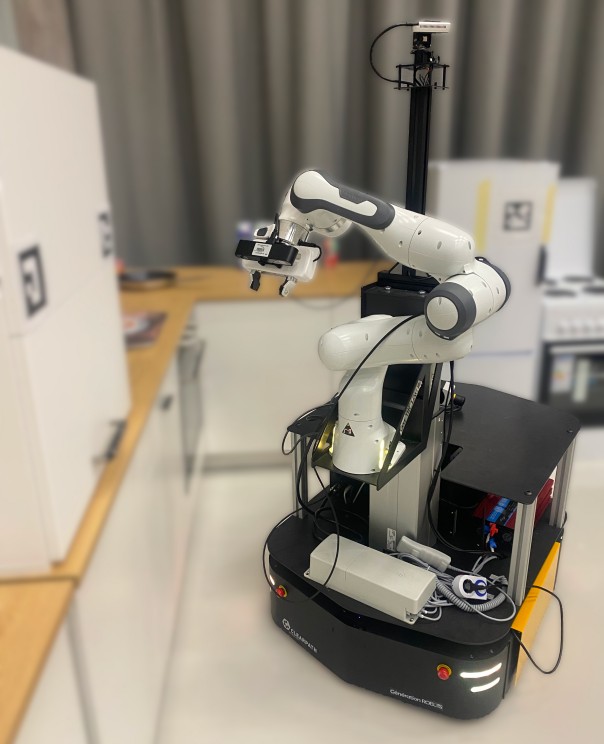}
    \caption{Robot setup showing the Franka manipulator, with an end-of-arm depth camera, mounted onto a mobile base.}
    \label{fig:robot_setup}
\end{figure}

We perform experiments using a Franka Panda robot arm mounted on a mobile robot base as shown in \cref{fig:robot_setup}. 
We consider a mixture of table-top manipulation tasks along with manipulation of articulated kitchen furniture, resulting in a total of 10 tasks. A full list of tasks is given in \cref{sec:appendix:task_description}, together with example images shown in \cref{fig:teaser} and \cref{fig:example_output}. For each task (except \texttt{plug\_charger}), we recorded five demonstrations with varying initial poses of both the manipulated objects and secondary objects.

In theory, demonstrations and inference can use RGB-D observations from different cameras, however, for convenience we both record demonstrations and run inference with a SteroLab ZED2i.
We record demonstration videos with a static camera position. This allows easy recording of human demonstrations from a third-person perspective and aligns with the prospective setting of having a robot watch a human demonstration and then being able to imitate it. 
For the purpose of faster computation, we subsample demonstration videos to a fixed length of $T=11$ observations, yielding ten steps, with a linear spacing between frames.

\subsection{Demo. Based Tracking \& Re-Detection Evaluation}
\label{subsec:exp:correspondence_tracking}

\input{assets/qualitative_experimental_results}
\input{assets/tables/offline_correspondences.tex}
\input{assets/tables/offline_relative_pose}
In the first offline experiment, we evaluate the task of finding correspondences in a target image given a source image and a mask from which we want to establish correspondences. This procedure is an integral part of and used twice in \ourName{}. Once, when tracking the object within a demonstration and once, when the object is re-detected in a live observation.
We evaluate the quality of the various correspondence methods outlined \cref{subsec:related_work:pose_estimation} and substantiate our decisions taken in \cref{sec:method}. 
We use the segmentation masks from our pre-processing procedure (refer to \cref{subsec:method:object_segmentation}) which includes manual verification of masks.
For the evaluation criterion, we count the percentage of correspondence points that remain within the segmentation masks in the next frame (precision) and their absolute amount (recall). 
We report the results in \cref{tab:tracking_eval}.

We observe that when establishing correspondences within a single demonstration (see \cref{subtab:tracking:intra_demo}) there is no significant precision increase of LoFTR ($89.4\%$) over RAFT ($88.2\%$) but RAFT's recall is $84\times$ higher. Using CNOS re-detection within a demonstration consistently decreases performance as the re-detection is more likely to fail due to the hand holding and thus, occluding the object heavily from time.
In contrast, when estimating correspondences between demonstrations (see \cref{subtab:tracking:inter_demo}), one observes that using the CNOS mask re-detection helps to overcome the shortcomings of the local flow estimation. On the other hand, the CNOS mask re-detection hurts the global LoFTR method, as it can generate incorrect mask crops and lead to premature failures.
Thus, we chose to track the object within a demonstration only using RAFT and between demonstrations LoFTR.

\subsection{Demo. Based Trajectory Generation Evaluation}
\label{subsec:exp:relative_pose_warping}
In this setup, we compare the relative change of poses between the transformed demonstration trajectory $\demoTrajectoryWarped$ and the pseudo ground truth trajectory $\liveTrajectory$. This comparison shows how well the relative pose estimation and mixing components perform.
As before, we compare against various combinations outlined in \cref{subsec:related_work:pose_estimation}.
Given two demonstration episodes from our set, we consider one of them $\demoRecording$ as our input to \ourName{} and the other one as our pseudo ground truth $\liveRecording$. Thus, we perform an offline evaluation under the assumption that the extracted trajectory $\liveTrajectory$ in $\liveRecording$ is valid. 
As verified in \cref{subsec:exp:correspondence_tracking} this is a reasonable assumption.
It is important to note that this experiment cannot yield perfect results unless the human performs the task in the exact same manner, which is nearly impossible.
To address these concerns, we took significant care to perform the same movement in all demonstrations.
We compare the difference in relative poses between trajectories for a given time step. We calculate the translation error through Euclidean distance and the rotation error using angle-axis.
Results are shown in ~\cref{tab:trajectory_prediction}.
As expected, no method achieves an error close to zero, but the overall results of this experiment align closely with those of in the inter-demonstration correspondence evaluation experiment (refer to \cref{subtab:tracking:inter_demo})

\subsection{Real Robot Evaluation}
\label{subsec:real_robot_evaluation}

Given the promising results of the previous experiments in \cref{subsec:exp:relative_pose_warping}, we evaluate \ourName{} on the real-world robot setup.
We additionally ablate using the proposed CNOS re-detection step (over LoFTR) to detect grasping regions as well as hand affordance.
Given our previously collected ten tasks, we again set these up under similar conditions. 
We then thoroughly evaluate \ourName{} on over 150 real-world executions and conclude the following results and drawbacks of our proposed method.
Overall, \ourName{} is able to correctly warp the demonstration trajectory to the live scene in $79\%$ of our evaluation runs. For the remaining runs \ourName{} fails because no correspondences could be established (e.g. when the objects are heavily rotated). In the case of successful transfers, the majority of subsequent failures are caused by the robot's kinematic constraints which are limited compared to the human teacher.
For a visualization of success and failure cases we refer to \cref{fig:example_output}. %

{\noindent \textbf{In-Depth Task Analysis}:}
For the easiest pick-and-place tasks, \texttt{sponge\_tray} and \texttt{coke\_tray}, we achieve a high success rate even under modifications of the scene (e.g. tray moved up). The more difficult pick-and-place tasks, which require greater precision, \texttt{tennisball} and \texttt{cleanup\_box}, are also executed well when given a similar setup as shown in the demonstration.
For the re-orientation task \texttt{paperroll}, we frequently observe an execution to almost until the end of the task, just before the robot needs to lower its wrist, at which point it moves into its joint limits.
For the precise insertion tasks, \texttt{hanoi\_ring} and \texttt{plug\_charger}, as well as the pouring task \texttt{pour\_cup}, despite actual imprecisions in object pose estimation the most common failure case is grasping the object. Our used task objects are quite small and have intricate features. 
For the articulated object manipulation tasks, \texttt{cabinet\_open} and \texttt{drawer\_open}, we encounter two main problems preventing successful execution, first, due to sensor noise the predicted grasps on the narrow handles are colliding with the environment and second, the inverse kinematics solution often leading the robot into a singularity.
This behavior is expected as prior work has previously shown that for articulated object manipulation in the wild, a mobile base is beneficial \cite{mittal2022articulated}.

{\noindent \textbf{Ablation Results}:}
We see no significant difference when using the hand affordance-based grasp filtering.
Nonetheless, we would expect the filtering to yield an improvement when focusing on tasks where the grasp pose is crucial for task success.
Similar to the quantitative evaluation in \cref{subsec:exp:correspondence_tracking}, using CNOS as a pre-detection method sometimes fails catastrophically when similar objects are present in the scene.

%% file: assets/qualitative_experimental_results.tex
\begin{figure*}[h!]
    \captionsetup[subfigure]{labelformat=empty}
    \centering
     \begin{subfigure}[t]{0.11\textwidth}
        \centering
        \includegraphics[width=\textwidth]{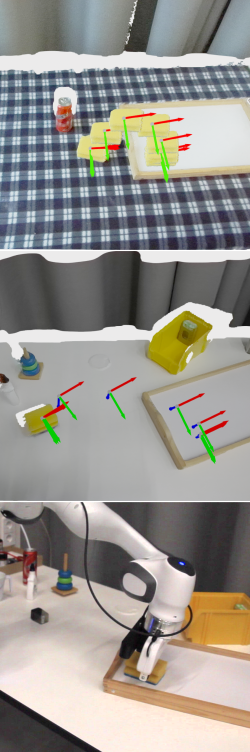}
        \caption{\texttt{sponge\_tray}}
        \label{subfig:eval:sponge_tray}
    \end{subfigure}
    \hfill
     \begin{subfigure}[t]{0.11\textwidth}
        \centering
        \includegraphics[width=\textwidth]{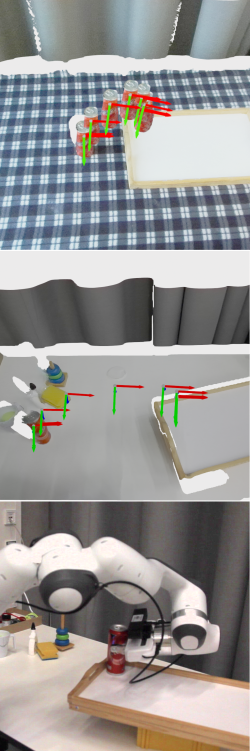}
        \caption{\texttt{coke\_tray}}
        \label{subfig:eval:coke_tray}
    \end{subfigure}
    \hfill
     \begin{subfigure}[t]{0.11\textwidth}
        \centering
        \includegraphics[width=\textwidth]{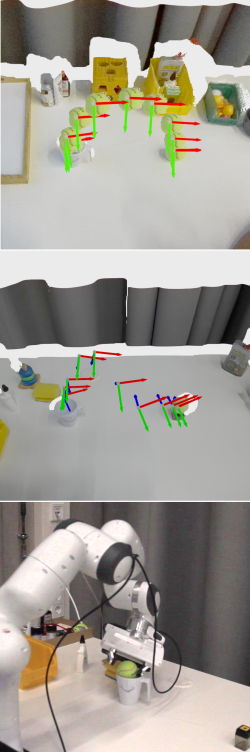}
        \caption{\texttt{tennisball}}
        \label{subfig:eval:tennisball}
    \end{subfigure}
    \hfill
     \begin{subfigure}[t]{0.11\textwidth}
        \centering
        \includegraphics[width=\textwidth]{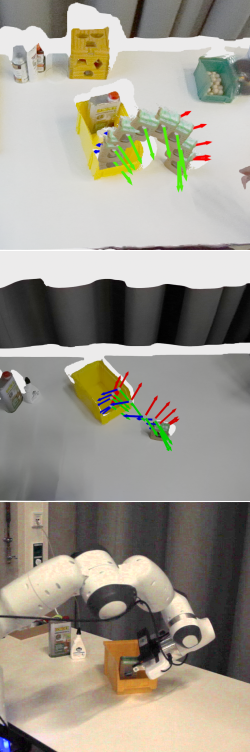}
        \caption{\texttt{cleanup\_box}}
        \label{subfig:eval:cleanupbox}
    \end{subfigure}
    \hfill
     \begin{subfigure}[t]{0.11\textwidth}
        \centering
        \includegraphics[width=\textwidth]{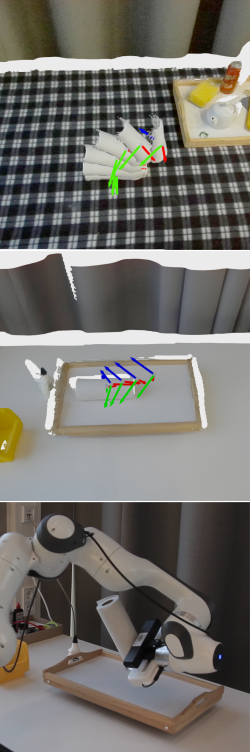}
        \caption{\texttt{paperroll}}
        \label{subfig:eval:paperroll}
    \end{subfigure}
    \hfill
     \begin{subfigure}[t]{0.11\textwidth}
        \centering
        \includegraphics[width=\textwidth]{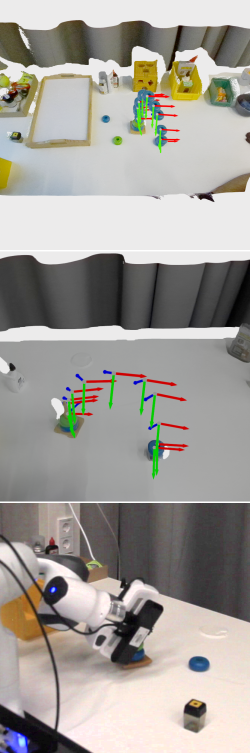}
        \caption{\texttt{hanoi\_tower}}
        \label{subfig:eval:hanoi_tower}
    \end{subfigure}
    \hfill
     \begin{subfigure}[t]{0.11\textwidth}
        \includegraphics[width=\textwidth]{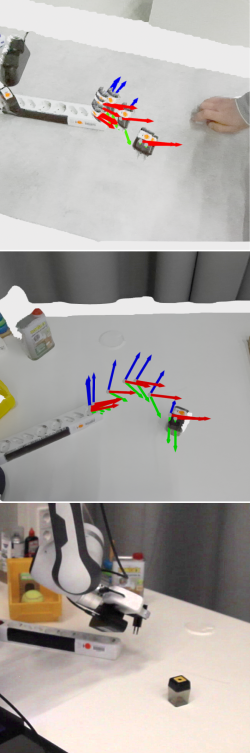}
        \caption{\texttt{plug\_charger}}
        \label{subfig:eval:plug_charger}
    \end{subfigure}
    \hfill
     \begin{subfigure}[t]{0.11\textwidth}
        \centering
        \includegraphics[width=\textwidth]{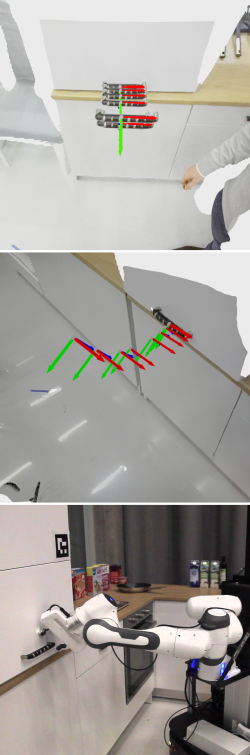}
        \caption{\texttt{cabinet\_open}}
        \label{subfig:eval:cabinet_open}
    \end{subfigure}
    \caption{Examples of trajectory generation, shown for various different tasks. 
    \textit{Top row}: rendered examples of trajectories extracted from human demonstrations, in-painted into the initial demonstration observation.
    \textit{Middle row}: rendered trajectories that have been generated in-situ for the robot imitation, in-painted into the live robot view.
    \textit{Bottom row}: images from live robot imitation runs.
    }
    
    \label{fig:example_output}
\end{figure*}

%% file: assets/tables/offline_correspondences.tex
\begin{table}[t]
    \centering
    \begin{subtable}{\linewidth}
        \centering
        \begin{tabular}{l l || c  c | c}
        \toprule
            \multicolumn{2}{c}{Method}                  & \multicolumn{2}{c|}{Segm. Inlier-} & Runtime\\
            Corresp.  & Detection & rate [$\%$]  & num. [N]  & [s]  \\
            \midrule
            RAFT \cite{Teed2020}     & -                     & 88.2    & \textbf{5472}   & \textbf{0.56}\\ %
            RAFT \cite{Teed2020}     & CNOS\cite{Nguyen2023} & 77.7    & 4821            & 8.18 \\ %
            LoFTR \cite{Sun2021}     & -                     & \textbf{89.4} & 65        & 0.80 \\ %
            LoFTR \cite{Sun2021}     & CNOS\cite{Nguyen2023} & 72.8    & 32              & 8.29\\ %
        \bottomrule
        \end{tabular}
        \caption{Tracking evaluation, within single demonstrations.}
        \label{subtab:tracking:intra_demo}
        \vspace{1.2em}
    \end{subtable}\\
    \begin{subtable}{\linewidth}
        \centering
        \begin{tabular}{l  l || c  c | c}
        \toprule
            \multicolumn{2}{c}{Method}                  & \multicolumn{2}{c|}{Segm. Inlier-} & Runtime\\
            Corresp.  & Detection & rate [$\%$]  & num. [N]  & [s]  \\
        \midrule
            RAFT \cite{Teed2020}     & -                     & 56.4 &         2308  &  \textbf{0.54} \\ %
            RAFT \cite{Teed2020}     & CNOS\cite{Nguyen2023} & 71.4 & \textbf{2922} &          7.99  \\ %
            LoFTR \cite{Sun2021}     & -             & \textbf{79.4}&           25  &          0.75 \\ %
            LoFTR \cite{Sun2021}     & CNOS\cite{Nguyen2023} & 63.4 &           14  &          8.02 \\ %

        \bottomrule
        \end{tabular}
        \caption{Re-detection evaluation, between different demonstrations.}
        \label{subtab:tracking:inter_demo}
    \end{subtable}
    \caption{Tracking and re-detection rvaluation based on correspondence methods. We evaluate the performance of correspondence methods by detecting correspondences between a source and a target image given a source mask. We measure the absolute inlier count~($N$) of established correspondences as well as the percentage of inliers~($\%$) that map to the ground truth target mask. We evaluate on two different setups, the first one, being within a demonstration (in \cref{subtab:tracking:intra_demo}) and the second one, between the initial observation of two demonstrations (in \cref{subtab:tracking:inter_demo}). 
    }
    \label{tab:tracking_eval} 
\end{table}

%% file: assets/tables/offline_relative_pose.tex
\begin{table}[t]
    \centering
    \resizebox{.8\columnwidth}{!}{
    \begin{tabular}{l l ||cc}
        \toprule
        \multicolumn{2}{c||}{Method} & \multicolumn{2}{c}{Traj. Pose Errors} \\
         Corresp. & Detection &  Rot. [rad] & Trns. [m]  \\
         \midrule
         RAFT \cite{Teed2020} & \tblPH    & 0.2243 & 0.0401 \\
         RAFT \cite{Teed2020} & CNOS\cite{Nguyen2023}       & 0.2288 & 0.0441 \\
         LoFTR \cite{Sun2021} & \tblPH & \textbf{0.2226} & \textbf{0.0387}\\
         LoFTR \cite{Sun2021} & CNOS\cite{Nguyen2023}      & 0.2417 & 0.0437 \\
         \bottomrule
    \end{tabular}
    }
    \caption{Relative Trajectory Pose Estimation Errors. For a set of demonstrations, we assume one of them as our given demonstration. From the set of remaining demonstrations, we will use the first image as a hypothesized live observation. We then calculate the translation and rotation error between the relative change in the demonstration trajectory and the generated trajectory. This assumes that each trajectory is executed with the same movement (direction and speed) and thus, the change between two steps should be similar. Note that this can only be considered a pseudo-error as the assumption can not be strictly enforced, due to human errors. 
    Pose estimation is done using least-squares rigid motion.
    }
    \label{tab:trajectory_prediction}
\end{table}

%% file: sections/05_conclusion.tex
\section{Conclusion}
We present \ourName{}, a modular method for strong one-shot imitation from human demonstration videos. We evaluated different variations of \ourName{} in an offline manner, proving its potential use in real-world robotic tasks as well as on a real robot setup, demonstrating its efficacy, and identifying key weaknesses. To facilitate, future research we made the code publicly available.
Potential improvements could include 
setting up a standardized benchmark that allows future researchers to 
iterate on subcomponents separately in order to improve the overall performance of \ourName{}. For instance, this could include fine-tuning the segmentation model or locally refining the predicted grasps.
The robotic execution can also be improved, for example, by integrating the robot's mobile base into the motion planning process to tackle tasks that are kinematically more challenging.

%% file: sections/06_appendix.tex
\subsection{Secondary Object Segmentation}
\label{sec:appendix:goal_heuristic}
In cases where Hands23~\cite{Shan2020} did not detect a secondary object mask, we propose an alternative approach. At the last time step of a manipulation action, we first use SAM~\cite{Kirillov2023} to fully segment the image. Given each mask, we will lift them into 3D, resulting in a set of point clouds. We then chose the point cloud (and consequently the secondary object mask) with smallest minimal distance to the manipulated object point cloud, i.e. if they are in contact the distance is 0.

\subsection{Combining Trajectories}
\label{sec:appendix:mixing}

Given two trajectories, at each timestep $t$ we interpolate between them using a time-dependent mixing weight $\interCoefficient(\timestep)$.
The positions are summed and the rotation is interpolated using slerp~\cite{Shoemake1985}.
The mixing weight 
\begin{equation}
    \interCoefficient(\timestep) = G \left( t~|~0,~\interCoefficientScaling (\allTimesteps-1) \right) \in \realSpace
\end{equation}
is a Gaussian distributed coefficient where $\interCoefficientScaling$ is a hyperparameter controlling the steepness of the mixing curve and $T-1$ is the number of trajectory steps.
If too steep, i.e. too small $\interCoefficientScaling \rightarrow 0$, there will be a sudden jump in the middle of the trajectory, if too flat, i.e. too large $\interCoefficientScaling \rightarrow \inf$, two sudden jumps will happen close to the beginning and the end. 
We chose $\sigma=1/2$ as it is a good trade-off between both.

\subsection{List of Experimental Tasks}
\label{sec:appendix:task_description}

A list of our experimental tasks with brief descriptions is detailed in \cref{tab:transposed_text}.

\input{assets/tables/task_list}

%% file: assets/tables/task_list.tex
\begin{table}[htb]
    \centering
    \resizebox{\columnwidth}{!}{%
    \begin{tabular}{llll}
    \hline
    \toprule
   {Name}                  & {Object}               & {Secondary Object}  & {Action Type} \\
    \midrule
    \texttt{coke\_tray}      & Coke can                & Kitchen tray   & Pick and place \\
    \texttt{sponge\_tray}    & Sponge                  & Kitchen tray   & Pick and place \\
    \texttt{drawer\_open}    & Drawer handle           & \textit{-}     & Articulated obj. man. \\
    \texttt{cabinet\_open}   & Cabinet handle          & \textit{-}     & Articulated obj. man.\\
    \texttt{hanoi\_ring}     & Hanoi tower ring        & Wood peg       & Insertion         \\ 
    \texttt{plug\_charger}   & Phone charger           & Socket bar     & Insertion         \\ 
    \texttt{pour\_cup}       & Mug                     & Gray bowl      & Pouring         \\ 
    \texttt{paperroll}      & Paper-towel roll        & -              & Re-orienting \\ 
    \texttt{tennisball}     & Tennis ball             & Cup            & Pick and place  \\ 
    \texttt{cleanup\_box}    & Cardboard box           & Storage box    & Pick and place  \\ 
    \bottomrule
    \end{tabular}%
    }%
    \caption{Overview of the tasks used in the experiments. Some example images are shown in \cref{fig:teaser}.}
    \label{tab:transposed_text}
\end{table}